\pdfoutput=1

\documentclass[11pt]{article}

\usepackage{EMNLP2022}

\usepackage{times}
\usepackage{latexsym}

\usepackage[T1]{fontenc}

\usepackage[utf8]{inputenc}

\usepackage{microtype}

\usepackage{inconsolata}

\usepackage{amsmath}  
\usepackage{xcolor}
\usepackage{amsthm}
\usepackage{xcolor,colortbl}
\usepackage{multirow}
\usepackage{empheq}
\usepackage{graphicx}
\usepackage{adjustbox}
\usepackage{booktabs}
\usepackage{wrapfig}
\usepackage{tabularx}

\DeclareMathOperator*{\sign}{\mathrm{sign}}
\DeclareMathOperator*{\inputembed}{{input\_embed}}
\DeclareMathOperator*{\red}{\mathrm{red}}
\DeclareMathOperator*{\tcross}{\mathrm{T}_\mathrm{cross}}
\DeclareMathOperator*{\scross}{\mathrm{s}_\mathrm{cross}}

\title{Proxy-based Zero-Shot Entity Linking by Effective Candidate Retrieval}

\author{Maciej Wiatrak\textsuperscript{1}\thanks{\noindent\textsuperscript{} \,\,Corresponding author.}, Eirini Arvaniti\textsuperscript{1}, Angus Brayne\textsuperscript{1}, Jonas Vetterle\textsuperscript{1, 2}, Aaron Sim\textsuperscript{1} \\
  \textsuperscript{1}BenevolentAI \textsuperscript{2}Moonfire Ventures \\
  London, United Kingdom \\
  \texttt{\{maciej.wiatrak, eirini.arvaniti, angus.brayne, aaron.sim\}@benevolent.ai} \\
  \texttt{jonas@moonfire.com}}
  
\begin{document}
\maketitle
\begin{abstract}
A recent advancement in the domain of biomedical Entity Linking is the development of powerful two-stage algorithms – an initial \emph{candidate retrieval} stage that generates a shortlist of entities for each mention, followed by a \emph{candidate ranking} stage. However, the effectiveness of both stages are inextricably dependent on computationally expensive components. Specifically, in candidate retrieval via dense representation retrieval it is important to have \emph{hard} negative samples, which require repeated forward passes and nearest neighbour searches across the entire entity label set throughout training. In this work, we show that pairing a proxy-based metric learning loss with an adversarial regularizer provides an efficient alternative to hard negative sampling in the candidate retrieval stage. In particular, we show competitive performance on the recall@1 metric, thereby providing the option to leave out the expensive candidate ranking step. Finally, we demonstrate how the model can be used in a zero-shot setting to discover out of knowledge base biomedical entities. 

\end{abstract}

\section{Introduction}
\label{Introduction}


The defining challenge in biomedical Entity Linking (EL) is performing classification over a large number of entity labels with limited availability of labelled mention data, in a constantly evolving knowledge base. For instance, while the Unified Medical Language System (UMLS) knowledge base \cite{Bodenreider2004} contains millions of unique entity labels, the EL training data in the biomedical domain as a whole is notoriously scarce, particularly when compared to the general domain -- Wikipedia, for instance, is powerful as \emph{both} a Knowledge base and a source of matching entities and mentions. Furthermore, biomedical knowledge bases are evolving rapidly with new entities being added constantly. Given this knowledge base evolution and scarcity of training data it is crucial that biomedical entity linking systems can scale efficiently to large entity sets, and can discover or discern entities outside of the knowledge base and training data.

Recent methods in the general entity linking domain \cite{logeswaran-etal-2019-zero, wu-etal-2020-scalable} address the data issue with zero-shot entity linking systems that use entity descriptions to form entity representations and generalise to entities without mentions. A particularly powerful architecture was initially proposed by \citet{humeau2020polyencoders} and further improved by \citet{wu-etal-2020-scalable}. It consists of a two-stage approach: 1) candidate retrieval in a dense space performed by a \textit{bi-encoder} \cite{wu-etal-2020-scalable} which independently embeds the entity mention and its description, and 2) candidate ranking performed by a \textit{cross-encoder} which attends across both the mention and entity description \cite{logeswaran-etal-2019-zero}. In this work we focus on the former, which is traditionally optimised with the cross-entropy (CE) loss and aims to maximise the similarity between the entity mention and its description relative to the similarities of incorrect mention-description pairs. In practice, the large number of knowledge base entities necessitates the use of negative sampling to avoid the computational burden of comparing each mention to all of the entity descriptions. However, if the sampled distribution of negatives is not reflective of the model distribution, the performance may be poor. Recently, \citet{zhang-stratos-2021-understanding} showed that using hard negatives - the highest scoring incorrect examples - results in bias reduction through better approximation of the model distribution. Collecting hard negatives is computationally expensive, as it requires periodically performing inference and retrieving approximate nearest neighbours for each mention.

At the ranking stage, negative sampling is not required, as the number of candidates usually does not exceed 64. However, the state-of-the-art cross-encoder model used for ranking is very expensive to run, scaling quadratically with the input sequence length. This highlights the need for efficient and performant candidate retrieval models capable of disambiguating mentions without the need for the expensive ranking step. 

In this paper, we propose and evaluate a novel loss for the candidate retrieval model, which breaks the dependency between the positive and negative pairs. Our contributions are: (1) a novel loss which significantly outperforms the benchmark cross-entropy loss on the candidate retrieval task when using random negatives, and performs competitively when using hard negatives. (2) We design and apply an adversarial regularization method, based on the Fast Gradient Sign Method \cite{goodfellow2014explaining}, which is designed to simulate hard negative samples without expensively mining them. (3) We construct a biomedical dataset for out of knowledge base detection evaluation using the MedMentions corpus and show that our model can robustly identify mentions that lack a corresponding entry in the knowledge base, while maintaining high performance on the retrieval task.

Our main testing ground is the biomedical entity linking dataset MedMentions \cite{Mohan2019}, which utilizes UMLS as its knowledge base. Additionally, to confirm that our method works also in the general, non-biomedical domain, we evaluate it on the Zero-Shot Entity Linking (ZESHEL) dataset proposed in \citet{logeswaran-etal-2019-zero}. We focus on the retrieval task with the recall@1 metric, because we are aiming to predict the entity directly without requiring the additional expensive ranking stage. Our results show that both the proposed loss and regularization improve performance, achieving state-of-the-art results on recall@1 and competitive performance on recall@64 on both datasets. Finally, we demonstrate that our model can robustly identify biomedical out of knowledge base entities, without requiring any changes to the training procedure.

\section{Related Work}
\label{Related Work}

\paragraph{Zero-Shot Entity Linking} There is a plethora of work on zero-shot entity linking methods leveraging the bi-encoder architecture \cite{wu-etal-2020-scalable} for candidate retrieval. These include novel scoring functions between the input and the label \cite{humeau2020polyencoders, luan-etal-2021-sparse, khattabzaharia}, cross-domain pretraining methods \cite{varma-etal-2021-cross-domain}, training and inference optimisation techniques \cite{Bhowmik2021FastAE} and effective entity representation methods \cite{ma-etal-2021-muver}.  Our work instead focuses on optimising the candidate retriever's loss function. 

The impact of hard negatives on the entity linking model performance has also been investigated \cite{Gillick2019, zhang-stratos-2021-understanding}. Notably, \citet{zhang-stratos-2021-understanding} develop analytical tools to explain the role of hard negatives and evaluate their model on the zero-shot entity linking task. We draw on this work, but move away from the CE loss towards a novel contrastive proxy-based loss.

Finally, there is a body of work on zero-shot entity linking in the biomedical domain using clustering \cite{angell-etal-2021-clustering, agarwal2021entity}. Our method does not consider the affinities between mentions directly and links them independently. Therefore, we do not study entity discovery.

An important aspect of biomedical entity linking systems is the detection of ``unlinkable'' mentions that lack a corresponding entry in the Knowledge Base - referred to as $\mathrm{NIL}$ detection. Methods for this task can be grouped into four main strategies \cite{shen2014entity, sevgili2020neural}: (1) label a mention as $\mathrm{NIL}$ when the corresponding candidate retriever does not return any candidate entities \cite{tsai2016cross}, (2) assign the $\mathrm{NIL}$ label to mentions whose corresponding top-ranked entity does not exceed some score threshold \cite{bunescu2006using, gottipati2011linking, lazic2015plato}, (3) train a classifier that predicts whether the top-ranked entity for a given mention is correct \cite{moreno2017combining}, (4) explicitly introduce a $\mathrm{NIL}$ class to the candidate ranking model \cite{kolitsas2018end}. A downside of the final approach is that knowledge of the $\mathrm{NIL}$ mention distribution is required at training time. In this work we tune a $\mathrm{NIL}$ score threshold (2) on a validation set. Detecting unlinkable mentions is particularly important in the biomedical domain, where the knowledge bases are rapidly evolving.

\paragraph{Proxy-based Losses} State-of-the-art entity linking models such as BLINK \cite{wu-etal-2020-scalable} leverage metric learning loss during training to make mentions similar to its assigned entity representations. Metric learning losses could be divided into two categories, pair-based and proxy-based losses \cite{Kim2020a}. Pair-based losses can leverage semantic relations between data points, here mentions. However, training them can be highly computationally expensive. On the other hand, proxy-based losses are significantly less computationally complex. This is done by establishing a proxy for each class and trying to increase the similarity between data points and its assigned proxies. Therefore, avoiding comparing the mentions to each other in favour of comparing the mentions to their proxies. We draw heavily on proxy-based losses \cite{Movshovitz-Attias2017a, Kim2020a} from metric learning by treating entity descriptions as the proxies. We establish a proxy for each entity, creating mention-proxy (i.e. entity) pairs, and optimise the model to embed the mention close to its assigned proxy. The loss proposed here is similar to the Proxy-NCA loss of \citet{Movshovitz-Attias2017a}. Our modification is the use of the Softplus function, similar to \citet{Kim2020a}, to avoid a vanishing gradient for the true mention-proxy pair.

\paragraph{Adversarial Regularization} Entity linking systems often rely on careful mining of hard negative examples to boost their performance \cite{Gillick2019, zhang-stratos-2021-understanding} at the expense of increased computational complexity. The model needs update hard negatives for each mention periodically. A potential alternative to hard negative mining is training on adversarial examples \cite{szegedy2013intriguing, goodfellow2014explaining} - synthetic data points designed to induce the model to making incorrect predictions, such that they are more challenging. Adversarial training can be seen as data augmentation and can help reduce overfitting. \citet{goodfellow2014explaining} introduced a simple method for generating adversarial examples, called Fast Gradient Sign Method (FGSM), which we build upon in this work. FGSM creates adversarial examples by applying small perturbations to the original inputs - often the word embeddings for NLP problems. FGSM has been used successfully as a regulariser in supervised and semi-supervised NLP tasks \cite{miyato2016adversarial, pan2021improved}. Here, we follow a similar approach and use FGSM to augment our training pairs with adversarial positive and negative examples.

\section{Task formulation}
\label{Task Formulation}

In the \textbf{Entity Linking task} we are provided with a list of documents $D \in \mathcal{D}$, where each document has a set of mentions $M_{D} = \{m_{1}, m_{2}, \dotsc, m_{N_D}\}$. The task is to link each mention $m_{i}$ to an entity $e_{i}$, where each entity belongs to the Knowledge Base (KB) $\mathcal{E}$. In this work we focus specifically on the problem of biomedical zero-shot entity linking. The setup for the zero-shot task is the same as for entity linking introduced above, except that the set of entities present in the test set is not present in the training set, i.e. $\mathcal{E}_{\mathrm{train}} \cap \mathcal{E}_{\mathrm{test}} = \emptyset$ with $\mathcal{E}_{\mathrm{train}} \cup \mathcal{E}_{\mathrm{test}} = \mathcal{E}$. We focus specifically on the \textbf{Candidate Retrieval} task, where the goal is given a mention $m_{i}$, reduce the pool of potential candidate entities from a KB to a smaller subset. Candidate retrieval is crucial for biomedical entity linking because of the large size of knowledge bases. In this work we use the bi-encoder architecture for candidate retrieval. Finally, in addition to the \textit{in-KB} entity linking task, where you only consider entities inside the KB, we also consider an \textbf{out of KB} scenario, where the task is to map mentions to the augmented set of labels $\mathcal{E} \cup \mathrm{NIL}$, with $\mathrm{NIL}$ indicating the absence of a corresponding KB entity.

\section{Methods}
\label{Methods}

\begin{figure}[ht!] 
    \centering
  \includegraphics[width=\columnwidth]{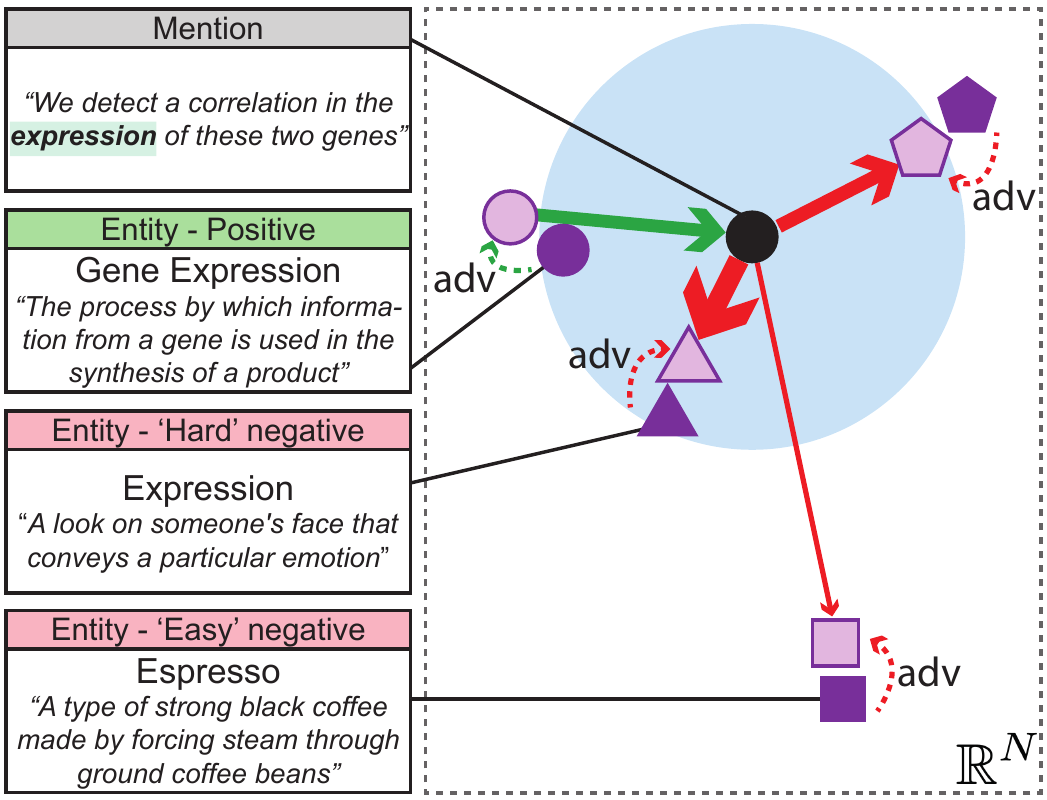}
  \caption{Overview of our proxy-based entity linking method. The mention and entity embeddings are encoded into a joint embedding space. During training, the magnitude of the gradients of the Proxy loss function with respect to the embedding coordinates is a function of the similarity between the mention and the entities (proxies). The gradients are represented by arrows whose widths indicate their magnitude. The $\mathrm{adv}$-labelled dotted arrows are the Fast Gradient Sign Method adversarial perturbations. The blue circle symbolizes the margin $\delta$.}
  \label{illustrated_method}
\end{figure}

In this section, we review the categorical CE loss, used by current state-of-the-art models, in the context of entity linking \cite{wu-etal-2020-scalable, zhang-stratos-2021-understanding}. We then compare it to our proposed Proxy-based loss. Finally, we describe and motivate our regularization approach.

\subsection{Loss} \label{loss}

Given a set of data points corresponding to mention representations $m \in M$ and to a set of proxies corresponding to entities $e \in \mathcal{E}$, the categorical CE loss is defined as:
\begin{equation}\label{firstlce}
    L_{\text{CE}}(m, P) := - \log\bigg(\frac{\exp(s(m, e^+))}{\sum_{e \in P} \exp(s(m, e))}\bigg),
\end{equation}
where $s$(·, ·) denotes a similarity function (e.g. cosine similarity or dot product), $e^{+}$ is the positive proxy for mention representation $m$, $P^{-}$ is a set of negative proxies used as negative samples, and $P = \{e^{+}\} \cup P^{-}$.

The gradient of the CE loss with respect to $s(m, e)$ is given by: 

\begin{equation}\label{grad_ce}
\frac{\partial L_{\text{CE}}}{\partial s(m, e)} = 
\begin{dcases}
      -1 + \frac{\exp(s(m, e^+))}{\sum_{e \in P} \exp(s(m, e))} , & e = e^{+} \\
      \frac{\exp(s(m, e^-))}{\sum_{e \in P} \exp(s(m, e))}, &  e \in P^{-}
\end{dcases}
\end{equation}

In practice training is performed with negative sampling. If the negatives are sampled randomly, often the exponential term for the positive entity is much larger than that of the negative samples and the gradients vanish. When $ s(m, e^+) \gg s(m, e^-) \ \forall e^- \in P^{-}$ then $ \partial L_{\text{CE}}/{\partial s(m, e)} \rightarrow 0 $. This behaviour is desirable when training with the full distribution of negative pairs, but stifles learning in the noisier sampling setup. A common approach is the use of hard negatives \cite{Gillick2019, zhang-stratos-2021-understanding}, which increases performance over training with random negatives at the cost of increased computational complexity. 

On the other hand, contrastive metric learning losses \cite{bromley1993, chopra2005, hadsell2006dimensionality} alleviate the vanishing gradients problem by decoupling the positive and negative loss terms. Proxy-based contrastive losses, such as Proxy-NCA \cite{Movshovitz-Attias2017a}, aim to increase the similarity between a data point $x$ and its assigned proxy $e^{+}$, while decreasing the similarity between $x$ and its negative proxies $e^{-} \in P^{-}$. As demonstrated in \cite{Kim2020a}, a downside of Proxy-NCA is that the scale of its gradient is constant for positive samples. This issue is alleviated by the Proxy Anchor loss \cite{Kim2020a}, whose gradient reflects the relative hardness of both positive and negative pairs, resulting in improved model performance. 

Drawing inspiration from the proxy-based metric learning losses described above, we formulate our Proxy-based (Pb) candidate retrieval loss as follows: 

\begin{equation}
\label{proxy_anchor_loss_eq}
\begin{split}
    L_{\mathrm{Pb}}(m, P) = \log(1 + \exp(-\alpha(s(m, e^{+}) - \delta)) \\ + \log(1 + \sum_{e^{-} \in P^{-}}\exp(\alpha(s(m, e^{-}) + \delta)),
\end{split}
\end{equation}

where we use the same notation as in Eq.~\ref{firstlce}. In addition, $\alpha$ is a hyperparameter controlling how strongly positive and negative samples pull and push each other, and $\delta$ is a margin. If $\alpha$ and $\delta$ are large, the model will be strongly penalized for the positive pair being too far from each other, and conversely the negative pair for being too close to each other. If $\alpha$ and $\delta$ are small, the model will receive weaker feedback. The Softplus function, a smooth approximation of the $\mathrm{ReLU}$, introduces an additional margin beyond which the model stops penalising both positive and negative pairs, thus reducing overfitting. The gradient of our Proxy-based loss function is given by:

\begin{equation}\label{grad_sp}
\frac{\partial L_{\text{Pb}}}{\partial s(m, e)} = 
\begin{dcases}
      \frac{-\alpha \exp(-\alpha s^{+})}{1 + \exp(-\alpha s^{+})} , & e = e^{+} \\
      \frac{\alpha \exp(\alpha s^{-})}{1 + \sum\limits_{e^{-} \in P^{-}} \exp(\alpha s{-})}, &  e \in P^{-}
\end{dcases}
\end{equation}

where $s^{+} = s(m, e^+) - \delta $, $s^{-} = s(m, e^-) + \delta $. This gradient reflects the relative hardness of negative examples, decoupled from the positive pair, which makes it less sensitive to the choice of negative sampling scheme.

\subsection{Regularization}\label{sec:methods-reg}
Our regularization approach is based on a simple adversarial training technique, called \textit{Fast Gradient Sign Method} (FGSM) \cite{goodfellow2014explaining}. The idea of FGSM is to generate adversarial examples according to the following equation:

\begin{equation}
x_{\mathrm{adv}} = x + \epsilon * \sign(\nabla_{x} L(x, y))
\end{equation}

where $x$ is the original training example, $y$ its corresponding label, $L$ the loss function that is minimised during model training, and $\epsilon$ a small number defining the magnitude of the perturbation.

FGSM applies a small perturbation to the input example that should not change the label of the resulting example $x_{\mathrm{adv}}$. However, \citet{goodfellow2014explaining} demonstrated that even infinitesimal perturbations can cause drastic changes to the model output when carefully designed. This effect is due to the locally linear nature of neural networks in combination with the high dimensionality of their inputs. Moreover, it is the direction, rather than the magnitude, of the perturbation that matters the most. In FGSM the direction is determined by the gradient of the loss function with respect to the model input - $x$ is pushed in the direction of highest loss increase given its true label $y$.

In the context of entity linking task, we are interested in generating examples adversarial to the learned metric, in other words hard negative and hard positive examples for a given mention $m$. To this end, we applied the following perturbations to the entity encoder input embeddings $z = \inputembed(e)$:
\begin{align}
z_{\mathrm{adv}}^{-} &= z^{-} + \epsilon * \sign(\nabla_{z^{-}} s(m, e^{-})) \label{eq:adv-neg} \\
z_{\mathrm{adv}}^{+} &= z^{+} - \epsilon * \sign(\nabla_{z^{+}} s(m, e^{+})) \label{eq:adv-pos}
\end{align}
where $m$ is the anchor mention and $z^{-}, z^{+}$ are the encoder input embeddings of negative and positive entities $e^{-}, e^{+}$ correspondingly.

Given $N$ negative entities for a mention $m$, the generated adversarial entity embeddings $P_{\mathrm{adv}} = \{ z_{\mathrm{adv\_1}}^{-}, \ldots, z_{\mathrm{adv\_N}}^{-}, z_{\mathrm{adv}}^{+} \}$ are used as additional training examples, giving rise to an auxiliary loss term that encourages the model to be invariant to local adversarial perturbations. Thus, the final objective we are trying to minimise becomes: 
\begin{equation}
L_{\mathrm{Pb}}(m, P) + \lambda L_{\mathrm{Pb}}(m, P_{\mathrm{adv}})
\end{equation}
where $\lambda$ is a hyperparameter controlling the relative contributions of the two losses.

\section{Experiments}
\label{Experiments}

\subsection{Datasets}

\paragraph{MedMentions} This is  is a biomedical entity-linking dataset consisting of over 4,000 PubMed abstracts \cite{Mohan2019}. As recommended by the authors, we use the ST21PV subset, which has around 200,000 mentions in total. A large number of mentions in both the validation and test splits are zero-shot, meaning their ground truth label is not present in the training data. We do not carry out any additional preprocessing on the dataset. Finally, for the knowledge base (KB), we follow the framework in \citet{varma-etal-2021-cross-domain} and use the UMLS 2017AA version filtered by the types present in the ST21PV subset. The final KB includes approximately 2.36M entities. 

To evaluate our models in the $\mathrm{NIL}$ detection setting, we have created a new dataset based on MedMentions. In this dataset, we have assigned mentions corresponding to 11 entity types a  $\mathrm{NIL}$ label and removed them from the Knowledge Base. Details on the dataset statistics and removed entity types can be found in the Appendix.

\begin{table}[]
\resizebox{\columnwidth}{!}{%
\begin{tabular}{lcccccc}
\hline
                 & \multicolumn{3}{c}{\textbf{MedMentions}}      & \multicolumn{3}{c}{\textbf{Zero-Shot EL}}     \\
                 & \textbf{Train} & \textbf{Val} & \textbf{Test} & \textbf{Train} & \textbf{Val} & \textbf{Test} \\ \hline
Mentions         & 120K           & 40K          & 40K           & 49K            & 10K          & 10K           \\
Entities         & 19K            & 8K           & 8K            & 333K           & 90K          & 70K           \\
\% Entities seen & 100            & 57.5         & 57.5          & 100            & 0            & 0             \\ \hline
\end{tabular}%
}
\caption{Statistics of datasets used. "\% Entities seen" signifies the percentage of ground truth entities seen during training.}
\label{tab:dataset-stats}
\end{table}

\paragraph{Zero-Shot Entity Linking dataset} ZESHEL, a general domain dataset was constructed by \citet{logeswaran-etal-2019-zero} from Wikias\footnote{https://wikia.com}. It consists of 16 independent Wikias. The task is to link mentions in each document to a Wikia-specific entity dictionary with provided entity descriptions. The dataset is zero-shot, meaning there is no overlap in entities between training, validation and test sets.

\subsection{Input Representation and Model Architecture}

Similarly to \citet{wu-etal-2020-scalable, zhang-stratos-2021-understanding, varma-etal-2021-cross-domain} our candidate retriever is a bi-encoder consisting of two independent BERT transformers. We use the bi-encoder to encode a textual mention and an entity description independently then obtain a similarity score between them. 

Namely, Given a mention and its surrounding context $\tau_{m}$ and an entity $\tau_{e}$, we obtain dense vector representations $\mathbf{y_{m}} = \red(T_{1}(\tau_{m}))$ and $\mathbf{y_{e}} = \red(T_{2}(\tau_{e}))$, where $T_{1}$ and $T_{2}$ are the two independent transformers of the bi-encoder and $\red(\cdot)$ is a function that reduces the output of a transformer into a single vector. We use a mean pooling operation for the function $\red(\cdot)$.


As in \citet{wu-etal-2020-scalable, zhang-stratos-2021-understanding, varma-etal-2021-cross-domain} we use the dot product to score the mention $\mathbf{y_{m}}$ against an entity vector $\mathbf{y_{e}}$ when using the CE loss. For our Proxy-based loss we use cosine similarity. 


In this, work, we focus on entity linking by efficient candidate retrieval, but we also include the ranker results using the highest scoring candidate entities in the Appendix, where we also include more details on entity, mention and context modelling.

\subsection{Training \& Evaluation Details}
In all our experiments we used the transformer architecture \cite{vaswani2017} for the encoders. Namely, we used BERT \cite{devlin-etal-2019-bert}, initialised with appropriate pre-trained weights: SapBERT \cite{liu-etal-2021-self} for MedMentions and the uncased BERT-base \cite{devlin-etal-2019-bert} for ZESHEL. For FGSM regularization, we apply adversarial perturbations to the composite token embeddings (i.e. sum of word, position and segment embeddings) used as input to BERT. We apply our regularization to both Proxy-based and CE. For information on hyperparameter tuning please refer to the Appendix. We tune all of our experiments on the validation set and report results on the test set. Due to hardware limitations, the training was conducted on a single V100 GPU machine with 16 GB of GPU memory. The limited GPU capacity, in particular, memory, posed a challenge by constraining us to using a relatively low number of negatives when training a retriever.

\subsubsection{Candidate Retriever}

The retriever model is optimised with the Proxy-based loss (\ref{proxy_anchor_loss_eq}) and benchmark CE loss (\ref{firstlce}) for fair comparison. We evaluate the retriever on the micro-averaged recall@1 and recall@64 metrics, where in our setup recall@1 is equivalent to accuracy. Here we focus on the recall@1 metric, which is highly relevant for efficient candidate retrieval models that do not necessitate  running an expensive cross-encoder for candidate ranking. We use two negative sampling techniques: (1) Random, where the negatives are sampled uniformly at random from all entities in the knowledge base, and (2) Mixed-p: p percent of the negatives are hard, the rest are random. This is  motivated by the results shown in Zhang and Stratos (2021). We set
the p to 50\%.

\paragraph{Hard negative mining} Retrieving hard negatives requires running the model in the inference mode over the entire KB. Then, for each mention, the most similar (i.e. hard) negatives are sampled according to a scoring function. Here, we use FAISS \cite{johnson2019billion} for obtaining hard negatives given a mention and an index of entity embeddings from the KB. 

Running a forward pass over the entire KB at regular intervals can be costly and time-consuming as the KB often amounts to millions of entities. Moreover, the computational complexity of retrieving hard negatives may grow exponentially depending on the scoring function. For example, the traditionally used scoring function also leveraged in this work, where the mention and entity are both represented with a single embedding requires $\mathcal{O}(me)$ approximate nearest neighbour searches, where $m$ and $e$ are the number of mentions and entities respectively. However, employing an alternative scoring function such as the \emph{sum-of-max} used in \citet{zhang-stratos-2021-understanding} which requires comparing a set of mention embeddings with a set of entity embeddings results in $\mathcal{O}(mexy)$ where $x$ and $y$ is the number of mention vector and entity vector embeddings. In \citet{zhang-stratos-2021-understanding} $x$ and $y$ are set to $128$, the number of maximum tokens in the mention and entity input sequence.

This highlights the computational cost of hard negative mining and underlines the need for both methods which can work effectively with random samples as well as more efficient hard negative mining strategies. In this work we propose a method for the former.

\paragraph{Biomedical Out of Knowledge Base Detection} For the biomedical $\mathrm{NIL}$ detection scenario training proceeds exactly as in the in-KB setting. We train models with the Proxy-based loss with different margins, and also a model with the CE loss. In each case, we use a validation set that includes $\mathrm{NIL}$ mentions to select an appropriate threshold for the retrieval model. Mentions whose corresponding top-ranked entity does not achieve this score are assigned the $\mathrm{NIL}$ label. We choose the threshold that maximises the F1 score for $\mathrm{NIL}$ entities in the validation set. We then apply this threshold to detect $\mathrm{NIL}$ mentions in the test set.

\begin{table}[ht!]
\resizebox{\columnwidth}{!}{%
\begin{tabular}{lccc}
\toprule
& \textbf{\# Neg.} & \textbf{recall@1} & \textbf{recall@64} \\ \cmidrule{1-4}
 \citet{angell-etal-2021-clustering} &  - & 50.8  & 85.3  \\
 \citet{agarwal2021entity} & - & 72.3  & 95.6  \\
 \citet{varma-etal-2021-cross-domain} & 100 & 71.7  & -  \\ \cmidrule{1-4}
\multirow{3}{*}{{\textbf{PEL-CE}}} & 32 (mixed)  & 72.1 & 95.5 \\
& 64 (mixed)  & 72.1 & 95.6 \\ 
 & 64 (random) & 55.7  & 94.0 \\ \cmidrule{2-4}
\multirow{3}{*}{{\textbf{PEL-Pb}}} & 32 (mixed)  & 71.6 & 93.3 \\
& 64 (mixed)  & \textbf{72.6} & 95.0 \\ 
& 64 (random) & 63.3 & \textbf{95.9} \\ \cmidrule{2-4}
\textbf{PEL-CE + FGSM} & 32 (mixed)   & 72.3 & 95.5 \\
\textbf{PEL-Pb + FGSM} & 32 (mixed)   & 72.4 & 93.7 \\
\bottomrule
\end{tabular}
}
\caption{Candidate retrieval results on the MedMentions dataset. CE and Pb refers to cross-entropy and proxy-based losses respectively. All experiments were run with mixed random and hard negatives ``(mixed)", or only ``(random)" negatives. The bold figures represent the best score for each recall metric. Note that FGSM PEL variants were only run with 32 negatives due to GPU memory constraints.}
\label{tab:medmentions-cand-gen}
\end{table}

\begin{table}[ht!]
\resizebox{\columnwidth}{!}{%
\begin{tabular}{lcccc}
\toprule
    & \multicolumn{2}{c}{\textbf{Random}} & \multicolumn{2}{c}{\textbf{Mixed}} \\
    &  recall@1 & recall@64 & recall@1 & recall@64 \\ \cmidrule{1-5}
\citet{Wu2019}$^\dag$ & - & 81.80  & 46.5  & 84.8  \\
\citet{agarwal2021entity} & 38.6  & 84.0  & 50.4 & 85.1  \\
\citet{ma-etal-2021-muver} & 45.4  & \textbf{90.8}  & -  & - \\
\citet{zhang-stratos-2021-understanding} & -  & 87.62  & -  & \cellcolor[HTML]{EFEFEF} \textbf{89.6}  \\ \cmidrule{1-5}
\textbf{PEL-CE} & 44.1  & 84.8  & 52.5  & 87.2 \\
\textbf{PEL-Pb} & 48.9  & 85.2  & 53.1 & 86.0 \\
\textbf{PEL-CE + FGSM} & 44.1  & 85.2  & 53.2  & 87.2  \\
\textbf{PEL-Pb + FGSM} & \textbf{49.7}  & 85.6  & \cellcolor[HTML]{EFEFEF} \textbf{54.2}  & 86.6  \\ 
\bottomrule
\end{tabular}
}
\caption{Candidate retrieval results on the ZESHEL dataset. CE and Pb refers to cross-entropy and proxy-based losses respectively. The negative to positive sample ratio for all PEL runs is 32. The bold figures represent the best score for each sampling strategy (random vs. mixed random and hard). The highlighted figure represents the best overall score across strategies. \dag we use the results reported in \citet{zhang-stratos-2021-understanding} for random negatives and \citet{ma-etal-2021-muver} for mixed negatives.
}
\label{tab:zeshel-cand-gen}
\end{table}

\section{Results}
\label{Results}

We present the results for candidate retrieval and benchmark our models against suitable methods. We name our method Proxy-based Entity Linking (PEL-Pb). We also report the results of a version of our model which uses the CE (PEL-CE) loss on all experiments for comparison.

\subsection{MedMentions}

Table \ref{tab:medmentions-cand-gen} shows that all approaches using bi-encoder transformer models strongly outperform the N-Gram TF-IDF proposed in \citet{angell-etal-2021-clustering} for recall@1 and also recall@64. We also observe the strong positive effect of including hard negatives during model training. The effect is particularly strong for the CE loss, where recall@1 increases by 17\% compared with training on random negatives. We believe that such difference is partly due to the large size of the KB MedMentions KB, amounting to 2.36M entities, which contributes to the importance of hard negative mining. For the Proxy-based loss, including hard negatives increases recall@1 by 9\%, achieving state-of-the-art performance of 72.6\%. Adding FGSM regularisation boosted performance, as can be seen from the experiments with 32 negatives (the largest number of negatives we could fit into GPU memory when applying FGSM). However, it did not exceed the performance of the unregularized model with 64 negative samples.

\begin{table}[h!]
\resizebox{\columnwidth}{!}{%
\begin{tabular}{lccccc}
\toprule
                 & \multicolumn{3}{c}{\textbf{$\mathrm{NIL}$}}      & \multicolumn{2}{c}{\textbf{All classes incl. $\mathrm{NIL}$}}     \\
                 & \textbf{auPR} & \textbf{Precision} & \textbf{Recall} & \textbf{Recall@1} & \textbf{Recall@64} \\ \cmidrule{1-6}
Pb (m=0)         & 83.7       & 81.2          & 71.0           & \textbf{72.6}    & \textbf{90.4}   \\
Pb (m=0.01)         & 84.4       & 81.6           & 71.5            & 72.5   & 90.2        \\
Pb (m=0.05)         & 85.8  & 83.3     & 73.5 & 72.4  & 89.9 \\
Pb (m=0.1)         & \textbf{87.6}  & \textbf{85.2}     & \textbf{79.2} & 69.4  & 85.7 \\
CE             & 32.3   & 31.8      & 74.0      & 64.4           & 76.1        \\
\bottomrule
\end{tabular}
}
\caption{$\mathrm{NIL}$ detection results on the MedMentions dataset. auPR, precision and recall are reported exclusively for the $\mathrm{NIL}$ class, whereas micro-averaged recall@1 and recall@64 are reported for all classes including $\mathrm{NIL}$. Pb: Proxy-based with margin $m$, CE: Cross-Entropy.}
\label{tab:okb-results}
\end{table}

\begin{figure}[h!] 
  \centering
  \includegraphics[width=\columnwidth]{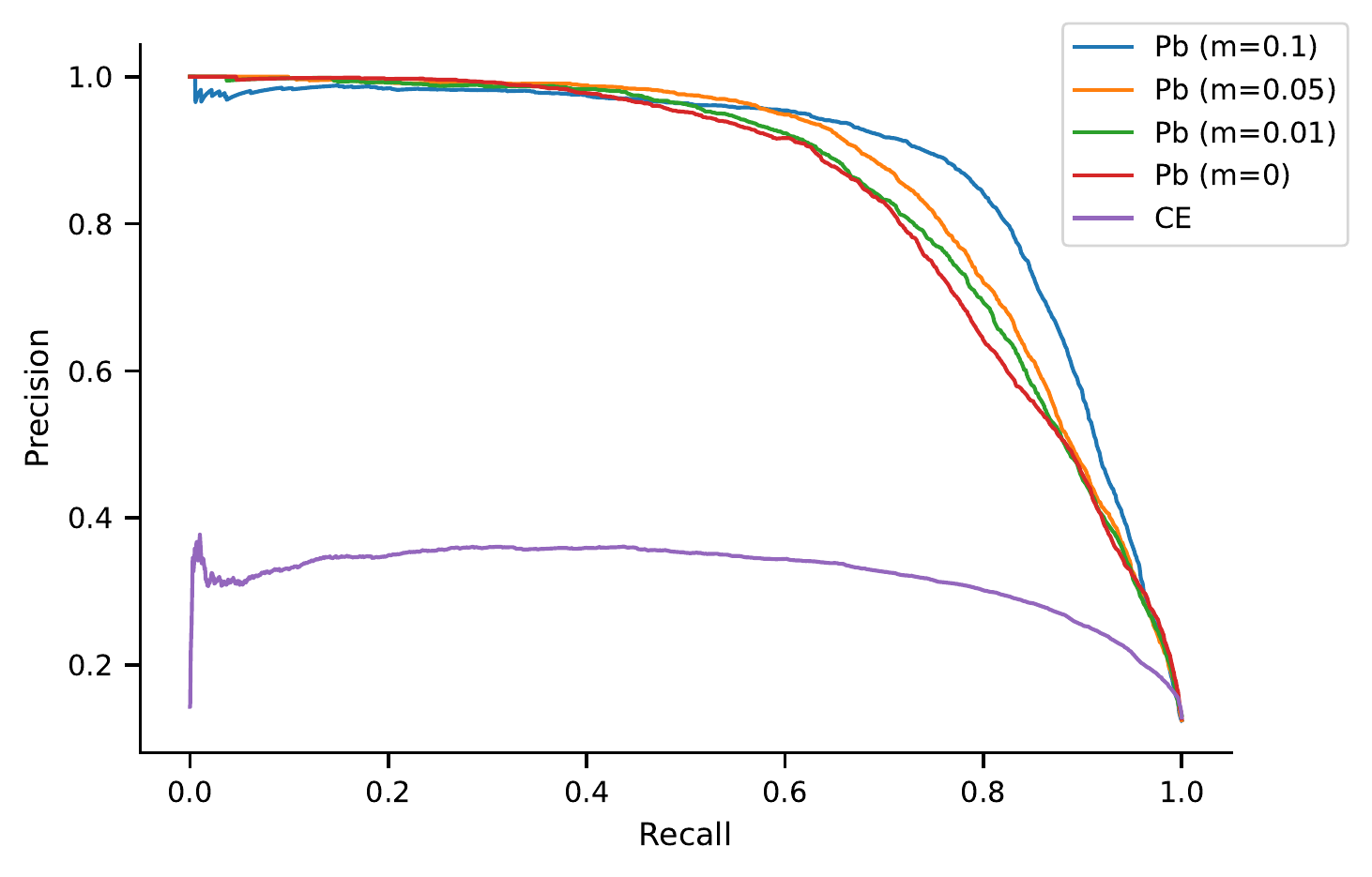}
  \caption{Precision Recall curves for $\mathrm{NIL}$ detection on the MedMentions dataset. Pb: Proxy-based with margin $m$, CE: Cross-Entropy.}
  \label{fig:okb_medmentions}
\end{figure}

\paragraph{Biomedical Out of Knowledge Base detection} We also evaluated our proposed loss function on $\mathrm{NIL}$ detection. All models trained with the Proxy-based loss significantly outperform the CE-based model in terms of both precision and recall (Figure \ref{fig:okb_medmentions}). The CE loss does not encourage low scores in absolute value for negatives examples, but rather encourages scores that are lower than the scores of positive examples. As we can see from the results, CE training fails to assign low scores to $\mathrm{NIL}$ mentions, as these are out-of-distribution negatives and thus have not been compared to positive examples during model training. Our Proxy-based loss does not suffer from this issue, even with a margin of $0$. We believe that this is accomplished by the decoupling of the positive and negative loss terms, such that low \textit{absolute} score values are encouraged for negative examples.

Furthermore, the higher the Proxy-based margin the better the model's performance with respect to detecting $\mathrm{NIL}$ mentions. At the same time, Proxy-based models with lower margins perform better at the overall recall metrics (Figure \ref{fig:okb_medmentions}). These metrics are computed with respect to all classes including the $\mathrm{NIL}$ class. Given that the performance differences among models with different margins are minimal, a practitioner could choose how to set the margin considering the trade-off between $\mathrm{NIL}$ detection and overall model performance. To our knowledge, we are the first to propose a method for $\mathrm{NIL}$ detection using the bi-encoder architecture.

\subsection{Zero-Shot Entity Linking dataset}

Based on the candidate retrieval results in Table \ref{tab:zeshel-cand-gen}, we can conclude six key points.
(1) Proxy-based models (Pb) outperform their Cross-Entropy (CE) counterparts across all considered settings for recall@1. In particular, our Proxy-based model using hard negatives and FGSM regularization achieves state-of-the-art recall@1 on this dataset. This highlights the gain that we get by breaking the dependency between positive and negative pairs.
(2) Including hard negatives always boosts model performance. This is particularly evident on the recall@1 metric. The model trained with CE loss strongly depends on hard negatives, with recall@1 increasing by 8\% compared to training with random negatives. For the Proxy-based loss the increase is 4\%, as the model already performs competitively when trained with random negatives. This showcases the importance of hard negative sampling for the CE loss. Hard negatives provide the model with much more meaningful feedback and avoid the threat of vanishing gradients (Eq.~\ref{grad_ce}).
(3) The difference between Pb and CE models becomes much smaller for recall@64. Trivially, as \textit{k} increases, recall@\textit{k} for all models will converge towards 1. Additionally, as \textit{k} increases to above the number of hard negatives, the model's ability to distinguish the hard negatives from the positive will not be seen in the metric.
(4) CE models marginally outperform Pb models with hard negatives at recall@64. Hard negatives consistently have a larger impact on CE compared to Pb also at recall@64 (2), while the benefits of Pb have been nullified as discussed in (3).
(5) Alternative methods leveraging the CE loss and different model architectures such as MuVER \cite{ma-etal-2021-muver} and SOM \cite{zhang-stratos-2021-understanding} outperform the bi-encoder based approach at recall@64. However, both MuVER and SOM are more complex models tuned for achieving high recall@64, whereas the main focus of our approach is high recall@1 in the pursuit of avoiding the additional ranking stage. Pb outperforms the only single stage entity linking model \citet{agarwal2021entity} across the board.
(6) FGSM regularization boosts the results of both Proxy-based and CE models, demonstrating its promise as a general method for regularizing the retrieval model.

\section{Discussion and Future Work}
\label{Discussion and Future Work}

We have proposed and evaluated a novel proxy-based loss for biomedical candidate retrieval.  Additionally, we have adopted an adversarial regularization technique designed to simulate hard negatives, and shown that both our loss and regularization boost performance on the recall@1 metric. We have also constructed a biomedical dataset for $\mathrm{NIL}$ detection and demonstrated that our candidate retrieval model can robustly identify biomedical $\mathrm{NIL}$ entities, while maintaining high overall performance. These are important advances towards closing the gap between the two-stage approach that include an expensive cross-encoder and a candidate retriever-only setup.

Notably, our work highlights the importance of hard negative sampling when optimising the candidate generator with the CE loss. Random negative sampling together with CE loss can result in the problem becoming trivial, for example the randomly sampled negative entity having a different type. However, accessing hard negative examples during model training can be challenging, particularly when the knowledge base is large and entity representations are frequently updated. 

Considering this, we recommend to employ our Proxy-based loss for the candidate retrieval task in three different scenarios: (1) training with random negatives, (2) optimising for recall@1, (3) detecting $\mathrm{NIL}$ entities. Moreover, we also recommend leveraging FGSM regularisation in any setup and both retrieval and ranking tasks.

An interesting approach would be to attempt to approximate hard negatives without frequent updates of the entity representations. This could potentially be done by keeping the entity encoder frozen, or exploring alternative relatedness measures which does not require frequently running the model over the whole knowledge base.  Finally, there is a plethora of work on proxy-based \cite{Movshovitz-Attias2017a, Kim2020a} and pair-based losses \cite{bromley1993, chopra2005, schroff2015, Dong_2018_ECCV}, usually discussed in the computer vision and metric learning literature. Improving the candidate retrieval is a crucial step towards high-performing and efficient entity linking systems that can be easily applied in real-world settings.

\section*{Limitations}

There are several limitations of our work. Firstly, we only demonstrate the advantages of our proposed method when computing hard negatives is computationally expensive, which is the case with large knowledge bases and expensive scoring methods. If computing hard negatives is not a bottleneck, one may use negative sampling with the baseline CE loss. However, biomedical knowledge bases typically contain a huge number of entities. Secondly, in our experiments we were limited to single GPU machines with at most 16GB of GPU memory. This prevented us from including more than 64 negatives samples in the standard setup and 32 negative samples when using FGSM regularization, which could potentially be benefit model performance. Thirdly, we acknowledge that some comparison to related work is missing, in particular, \citet{zhang-stratos-2021-understanding}. We were not able to reproduce the results cited in the paper using the publicly available code. Finally, our work is limited to proxy-based metric learning losses. More space could be devoted to the topic of how one could utilise metric learning more broadly for biomedical entity linking. We leave this for future work.

\section*{Ethics Statement}
The BERT-based models fine-tuned in this work and datasets are publicly available. We will also make our code as well as the biomedical out of knowledge base detection dataset publicly available.

The task of entity linking is often crucial for downstream applications, such as relation extraction, hence potential biases at the entity lining stage can have significant harmful downstream consequences. One source of such biases are the pre-trained language models fine-tuned in this work. There is a considerable body of work devoted to the topic of biases in language models. One way the entity linking systems can be particularly harmful is when they commit or propagate errors in the language models, knowledge bases, mention detection across certain populations such as races or genders. Because of the high ambiguity across biomedical mentions and entities in the knowledge base, it is important that the users investigate the output prediction of the entity linking system and often take is a suggestion, rather than gold standard. Finally, we highlight that linking the entity to its entry in the knowledge base and out of knowledge base detection can be analogous to surveillance and tracking in the computer vision domain, which comes with substantial ethical considerations.

\section*{Acknowledgements}
We thank Dane Corneil, Georgiana Neculae and Juha Iso-Sipil\"a for helpful feedbacks and the anonymous reviewers for constructive comments on the manuscript.

\bibliography{anthology,custom}
\bibliographystyle{acl_natbib}

\appendix

\section*{Appendices}

\begin{figure*}[htb!] 
  \centering
  \includegraphics[width=0.9\textwidth]{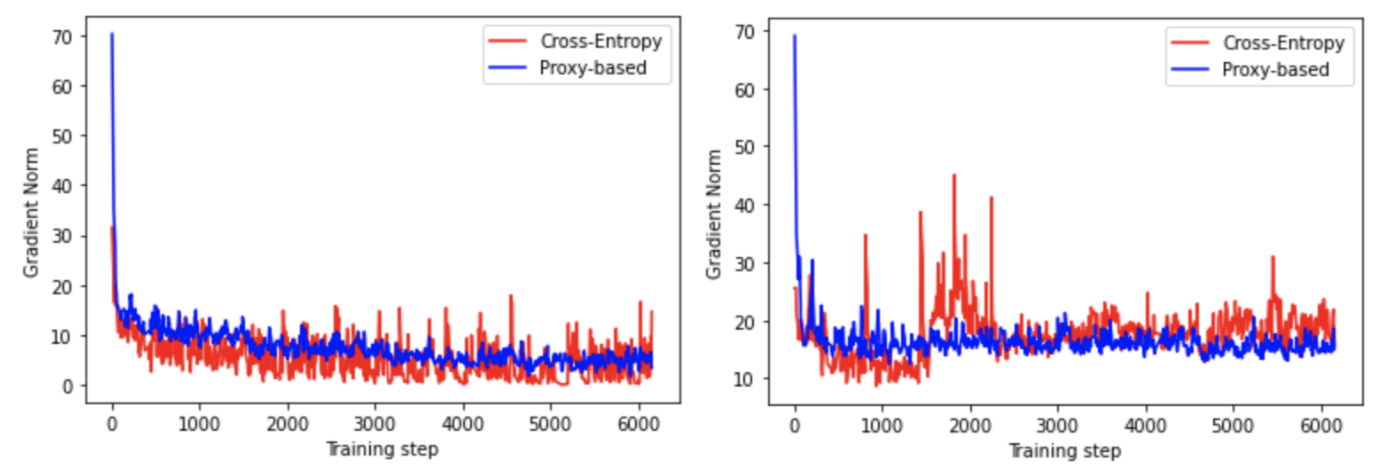}
  \caption{Comparison of smoothed gradient norms over training steps using two losses, CE and Proxy-based. The left plot visualizes the smoothed gradient norm when using random, and the right one leveraging mixed-50\% negatives. All the experiments were conducted on ZESHEL using 32 negatives.}
  \label{fig:grad_norms}
\end{figure*} 

\section{Context and Mention Modelling}
\label{sec:appendix}

We represent a mention and its surrounding context, $\tau_{m}$, as a sequence of word piece tokens
\begin{align*}
    [\mathrm{CLS}] \; \mathrm{ctxt_{l}} \; [\mathrm{M}_{s}] \; \mathrm{mention} \; [\mathrm{M}_{e}] \; \mathrm{ctxt_{r}} \; [\mathrm{SEP}]
\end{align*}
where $\mathrm{mention}$, $\mathrm{ctxt_{l}}$ and $\mathrm{ctxt_{r}}$ are the word-piece tokens of the mention, left and right context, and $[\mathrm{M}_{s}]$ and $[\mathrm{M}_{e}]$ are special tokens marking the start and end of a mention respectively. 

Due to the differences in available data, we represent entities differently for ZESHEL and MedMentions. On ZESHEL, we represent entities with a sequence of word piece tokens 
\begin{align*}
    [\mathrm{CLS}] \; \mathrm{title} \; [\mathrm{ENT}] \; \mathrm{description} \; [\mathrm{SEP}]
\end{align*}
where $[\mathrm{ENT}]$ is a special separator token. In contrast, when training on the MedMentions dataset we represent an entity by the sequence
\begin{align*}
    [\mathrm{CLS}] \; \mathrm{title} \; [\mathrm{SEP}] \; \mathrm{types} \; [\mathrm{SEP}] \; \mathrm{description} \; [\mathrm{SEP}]
\end{align*}
Descriptions of entities were sourced from UMLS.

\section{Candidate ranker setup and results}
\label{sec:appendix}

To evaluate the impact of our candidate retriever model on the downstream task of candidate ranking, we also conducted ranking experiments on both datasets.

\begin{table}[h!]
\centering
\small
\resizebox{\columnwidth}{!}{%
\begin{tabular}{clccc}
\toprule
& & \textbf{\# Candidates} & \textbf{Ranker} & \textbf{Accuracy} \\ \cmidrule{1-5}
\parbox[t]{2mm}{\multirow{6}{*}{\rotatebox[origin=c]{90}{ZESHEL}}} & \citet{wu-etal-2020-scalable} & 64 & Base & 61.3 \\
& \citet{wu-etal-2020-scalable} & 64 & Large & 63.0 \\
& \citet{zhang-stratos-2021-understanding} & 64 & Base & 66.7 \\
& \citet{zhang-stratos-2021-understanding} & 64 & Large & \textbf{67.1} \\
& \textbf{PEL-Pb} & 16 & Base & 62.8 \\
& \textbf{PEL-Pb + FGSM} & 16 & Base & 64.6 \\
\cmidrule{1-5}
\parbox[t]{2mm}{\multirow{7}{*}{\rotatebox[origin=c]{90}{MedMentions}}} & \citet{Bhowmik2021FastAE}$^{\dag}$ & - & - & 68.4 \\
& \citet{angell-etal-2021-clustering} & - & - & 72.8 \\
& \citet{varma-etal-2021-cross-domain} & 10 & Base & \textbf{74.6} \\
& \textbf{PEL-Pb} & 16 & Base & 74.0 \\
& \textbf{PEL-Pb + FGSM} & 16 & Base & \textbf{74.6} \\
\cmidrule{2-5}
& \citet{angell-etal-2021-clustering} & \multirow{2}{*}{{-}} & \multirow{2}{*}{{-}} & \multirow{2}{*}{{74.1}} \\
& + post-processing &  &  &  \\
& \citet{varma-etal-2021-cross-domain} & \multirow{2}{*}{{10}} & \multirow{2}{*}{{Base}} & \multirow{2}{*}{{\textbf{74.8}}} \\
& + post-processing &  &  &  \\
\bottomrule
\end{tabular}
}%
\caption{Ranker results on the ZESHEL and MedMentions datasets. \dag uses the full MedMentions dataset, rather than the ST21PV subset used by other models reported in the table and recommended by MedMentions authors'.}
\label{tab:reranker}
\end{table}

\paragraph{Training \& Evaluation setup} Similarly as in related work \cite{logeswaran-etal-2019-zero, wu-etal-2020-scalable, zhang-stratos-2021-understanding}, the highest scoring candidate entities from the candidate retriever are passed to a ranker, which is a cross-encoder consisting of one BERT transformer. The cross-encoder \citet{logeswaran-etal-2019-zero} is used to select the best entity out of the candidate pool. It takes as input $\tau_{m,e}$, which is the concatenation of mention/context and entity representations $\tau_{m}$ and $\tau_{e}$. We then obtain a dense vector representation for a mention-entity pair $\mathbf{y_{m, e}} = \tcross(\tau_{m, e})$, where $\tcross(\tau_{m, e})$ is the BERT transformer of the cross-encoder and $\red(\cdot)$ is a mean pooling function that takes the mean over input tokens embeddings. Entity candidates are scored by applying a linear layer $\scross(m, e) = \mathbf{y_{m, e}} \mathbf{W}$.

We pick the best performing retrieval model on recall@16 and use it to retrieve top 16 candidate entities for each mention. As the number of candidate entities is relatively low, we do not perform negative sampling and optimise the cross-encoder with the CE loss~(Eq. \ref{firstlce}). We report the micro-averaged unnormalized accuracy on the MedMentions dataset and macro-averaged unnormalized accuracy on the ZESHEL dataset in line with the prior work \cite{zhang-stratos-2021-understanding, wu-etal-2020-scalable}. The results are shown in the Table \ref{tab:reranker}.

\paragraph{Results} In Table \ref{tab:reranker} we can observe the downstream effect of having a candidate generator model with high recall@1 performance. On ZESHEL, We can see that a cross-encoder trained with the top 16 candidates from our best performing candidate generator achieved higher accuracy than \citet{wu-etal-2020-scalable} who used the top 64 candidates. Moreover, similarly as with the candidate retrieval, FGSM boosts performance. For completeness, we have also included the state-of-the-art results from \citet{zhang-stratos-2021-understanding} who used 64 candidates and a larger BERT model in the cross-encoder. In our experiments we were limited to a single GPU with 16 GB memory which restricted us to a low number of maximum candidates, namely $16$. We strongly believe that including more candidates than 16 would boost the performance of our method.

On MedMentions a cross-encoder trained with the top 16 candidates from our best performing candidate generator model achieved a competitive accuracy of 74\%. The accuracy further increased to 74.6\%  when adding FGSM regularisation, coming close to the state-of-the-art performance of \citet{varma-etal-2021-cross-domain}, which includes additional post-processing.

\section{Training details}
\label{sec:appendix}

The hyperparameters used for conducting the experiments are visible in Table \ref{tab:hyperparams}. We use a single NVIDIA V100 GPU with 16 GB of GPU memory for all model trainings. 

\begin{table}[h!]
\centering
\resizebox{\columnwidth}{!}{
\begin{tabular}{|l|l|l|}
\hline
\multicolumn{1}{|c|}{\textbf{Param}} & \multicolumn{1}{c|}{\textbf{Bi-encoder}} & \multicolumn{1}{c|}{\textbf{Cross-Encoder}} \\ \hline
Input sequence length           & 128   & 256   \\ \hline
learning rate           & 1e-5   & 2e-5   \\ \hline
warmup proportion       & 0.25   & 0.2    \\ \hline
\textit{eps}            & 1e-6   & 1e-6   \\ \hline
gradient clipping value & 1.0    & 1.0    \\ \hline
effective batch size    & 32     & 4      \\ \hline
epochs                  & 7      & 5      \\ \hline
learning rate scheduler & linear & linear \\ \hline
optimiser               & AdamW  & AdamW  \\ \hline
$\alpha$                   & 32     & -      \\ \hline
$\delta$                  & 0.0    & -      \\ \hline
FGSM $\lambda$            & 1      &  1   \\ \hline
FGSM $\epsilon$           & 0.01   & 0.01   \\ \hline
\end{tabular}
}%
\caption{Learning parameters for the bi-encoder and cross-encoder.}
\label{tab:hyperparams}
\end{table}

\section{Biomedical Out of Knowledge Base dataset details}
\label{sec:appendix}

We constructed the OKB dataset by replacing the label of a set of mentions from the MedMentions corpus \cite{Mohan2019} with the $\mathrm{NIL}$ class. Namely we pick the mentions belonging to 11 types: \textit{Mental Process}, \textit{Health Care Related Organization}, \textit{Element Ion or Isotope}, \textit{Medical Device}, \textit{Health Care Activity}, \textit{Diagnostic Procedure}, \textit{Professional or Occupational Group}, \textit{Mental Process}, \textit{Laboratory Procedure}, \textit{Regulation or Law}, \textit{Organization}, \textit{Professional Society}. The final OKB subset includes approximately 24K mentions and 3K unique entities. 

To ensure that the OKB dataset does not suffer from easy inferences and allows us to evaluate model performance. We ensured that the zero-shot distribution of the OKB mentions and types across the train/validation/test split was in line with the zero-shot distribution of mentions and types in the whole dataset. Additionally, we verified that there is no significant overlap between mention surface forms across the splits. Moreover, we looked at the length of entity descriptions which are used to create entity representations checking that the OKB mentions entity representations statistics are similar to the statistics computed using the whole dataset.

\section{Gradient norm analysis}
\label{sec:appendix}

\begin{table}[h!]
\begin{tabular}{llll}
\hline
                 & Train & Dev  & Test \\ \hline
Mentions         & 14K   & 4.8K & 4.7K \\
Entities         & 2.2K     & 1.1K    & 1.1K    \\
\% Entities seen & 100   & 57.7 & 57.5 \\ \hline
\end{tabular}
\caption{Statistics of the OKB MedMentions subset.}
\label{tab:okb-medmentions}
\end{table}

Figure \ref{fig:grad_norms} shows the behaviour of the gradient $l2$ norm for both losses. We can see that for both random and mixed negatives, the norm of the Proxy-based loss has considerably lower variance. This is visible particularly when using the mixed negatives.

\end{document}